\documentclass{ieeeaccess}
\usepackage{cite}
\usepackage{amsmath,amssymb,amsfonts}
\usepackage{algorithmic}
\usepackage{algorithm}
\usepackage{graphicx}
\usepackage{textcomp}
\usepackage{multirow}
\usepackage[english]{babel}
\usepackage{float}
\usepackage{amsmath}
\usepackage{caption}
\usepackage{url}
\def\BibTeX{{\rm B\kern-.05em{\sc i\kern-.025em b}\kern-.08em
    T\kern-.1667em\lower.7ex\hbox{E}\kern-.125emX}}

\begin{document}
\history{Date of publication xxxx 00, 0000, date of current version xxxx 00, 0000.}
\doi{10.1109/ACCESS.2017.DOI}

\title{Saliency-Driven Active Contour Model for Image Segmentation}
\author{
\uppercase{Ehtesham Iqbal}\authorrefmark{1},
\uppercase{Asim Niaz}\authorrefmark{2},
\uppercase{Asif Aziz Memon}\authorrefmark{1},
\uppercase{Usman Asim}\authorrefmark{1}
and KWANG NAM CHOI \authorrefmark{1}}

\address[1]{Department of Computer Science and Engineering, Chung-Ang University, Seoul 06974, South Korea}
\address[2]{Team STARS, Centre INRIA Sophia Antipolis, 06902 Valbonne, France}

\tfootnote {This work was supported by the National Research Foundation of Korea (NRF) grant funded by the Korea government Ministry of Science
and Information Technology (MSIT) (No. 2019R1F1A1062612). This research was supported by the Chung-Ang University Young Scientist Scholarship 2020.}

\markboth
{Author \headeretal: Preparation of Papers for IEEE TRANSACTIONS and JOURNALS}
{Author \headeretal: Preparation of Papers for IEEE TRANSACTIONS and JOURNALS}

\corresp{Corresponding author: Kwang Nam Choi (knchoi@cau.ac.kr)}

\begin{abstract}
Active contour models have achieved prominent success in the area of image segmentation, allowing complex objects to be segmented from the background for further analysis. Existing models can be divided into region-based active contour models and edge-based active contour models. However, both models use direct image data to achieve segmentation and face many challenging problems in terms of the initial contour position, noise sensitivity, local minima and inefficiency owing to the in-homogeneity of image intensities. The saliency map of an image changes the image representation, making it more visual and meaningful. In this study, we propose a novel model that uses the advantages of a saliency map with local image information (LIF) and overcomes the drawbacks of previous models. The proposed model is driven by a saliency map of an image and the local image information to enhance the progress of the active contour models. In this model, the saliency map of an image is first computed to find the saliency driven local fitting energy. Then, the saliency-driven local fitting energy is combined with the LIF model, resulting in a final novel energy functional. This final energy functional is formulated through a level set formulation, and regulation terms are added to evolve the contour more precisely across the object boundaries. The quality of the proposed method was verified on different synthetic images, real images and publicly available datasets, including medical images. The image segmentation results, and quantitative comparisons confirmed the contour initialization independence, noise insensitivity, and superior segmentation accuracy of the proposed model in comparison to the other segmentation models.
\end{abstract}


\titlepgskip=-15pt

\maketitle

\section{Introduction}
Image segmentation is often used to extract the informative parts from an image that are used for further analysis or understanding. This is an essential tool in computer image analysis, computer vision, medical imaging, image identification, and image classification [1-6]. In the past decades, numerous models have been proposed for image segmentation, including thresholding [7–9], clustering [10, 11], and active contour models (ACMs) [12, 13]. Active contour models are a productive approach for image segmentation, and there are two official types: parametric active contour (e.g, a snake [13]) and geometric active contour (level set based [12]) models. Parametric active contour models are highly dependent on the initial contour position; however, they cannot accurately handle an instant change in the curve topology, thereby limiting their application. By contrast, the curve movement of the geometric active contour model is dependent on the geometric 
 parameters rather than the expression parameters. Hence, these models can better deal with an instant change in the shape of the curves and are able to increase the range of the application. Compared with the available models, geometric active contour models that depend on the level set formulation [14] are more functional and effective. Images are segmented with respect to information such as color intensities and textures. Depending on the image information, we can further classify these models into edge-based ACMs [15-19] and region-based ACMs[20-24].
 
Edge-based models [25], as the name indicates, are dependent on the edges of an object to segment. An image gradient factor is used to find the edges of the object, and a larger image gradient shows the strong boundaries of the object. However, these models cannot detect an object with weak boundaries and edges. Moreover, the noise factor effects the gradient value, and the model represents false boundaries of the object. Therefore, this model cannot achieve satisfactory results for weak boundaries or noisy images.

By contrast, region-based ACMs do not depend on the image gradient and can segment images with weak boundaries. Region-based ACMs are constructed by internal and external energies applied to evolve the contour toward the boundaries of the target body. In this way, they have achieved better segmentation results for images with weak boundaries as compared to edge-based ACMs. For instance, the Chan and Vese (CV) model [12] is one of the best region-based ACMs available, having its roots in the Mumford-Shah (MS) model [26]. The CV model uses the global information of an image to find an energy functional that moves the contour toward the object boundaries and achieves better segmentation results for homogeneous images. However, images having a corrupted intensity or images with intensity inhomogeneity cannot be segmented by models using only the global image information energy.

To overcome the limitation of segmenting images with intensity inhomogeneity, Li et al. [27-28] proposed a local binary fitting (LBF) model using local image information to form a local fitting energy with a Gaussian kernel function. This model can segment images with intensity inhomogeneity but require four convolution operations during every iteration, which increases the computational cost. To overcome this limitation, Zhang et al. [29] proposed a local image fitting (LIF) energy functional model based on the concept of minimizing the difference between the fitted image and the original image. The LIF model achieves the same result as the LBF model but requires less time and fewer computations. Furthermore, Ding et al. [30] proposed a model that uses an optimized Laplacian of Gaussian energy with region-scalable fitting energy to segment the required area from the images. In addition, Xie et al. [31] detected ship targets in inshore SAR images with land architectures by introducing an improved level set method. An edge entropy dependent active contour model such as the edge entropy fitting energy [32] was also presented to enhance the results of the segmentation. Furthermore, some hybrid models[33-35] were also proposed for the segmentation of the images.

In computer vision, a saliency map of an image shows the unique quality of each pixel. Using a saliency map, we can simplify or change the representation of an image to make it more meaningful and easier to analyze. In a computer vision system or image processing, visual selective attention distinguishes the foreground and represents it prominently from the background. A saliency map extracts the visual attention of the image, and we can use it in an active contour model to obtain better image segmentation results. For example, Bai and Wang introduced the Saliency-SVM model in which they segment images into a binary classification [36] using saliency as the energy term. By contrast, Qin et al. merge the region saliency with an affinity propagation clustering algorithm and use a random walk method for segmentation [37]. In addition, Saliency-driven region edge based top down level set evolution model (SDREL) [38] uses a saliency map with a global active contour model for the segmentation of grayscale and color images. The segmentation results of SDREL are quite impressive, and it takes less time to evolve the contour toward the boundaries of the objects. SDREL uses a saliency map with a global image information based active contour model, and will therefore not produce prominent results for images with intensity inhomogeneity. Similarly, Nguyen et al. [39] and Yang et al. [40] also used visual saliency with active contour models to enhance the segmentation results. All such saliency-based active contour models can segment images that have strong boundaries and intensity homogeneity but fail to segment images with corrupted intensity and intensity inhomogeneity.

To obtain a better saliency driven image segmentation approach, inspired by the above-mentioned advantages and disadvantages of the existing active contour models, we derived a novel model, i.e., active contour control through visual saliency local fitting energy for image segmentation, using the advantages of LIF model and a saliency map. In the proposed model, we first find the saliency map of the original image, which is sensitive to the human visual system; then, we generate an energy function called local saliency fitting (LSF) energy by embedding saliency into the LIF model. Using LSF energy with LIF energy, a novel energy functional is proposed to obtain an improved saliency-based active contour model. Inspired by Li. et al. [27, 28], to obtain a stable evolution of the level set function $\phi$ and regularize the zero-level contour of $\phi$, we added a distance regularize term and the length of the zero level curve, respectively. Extensive experiments conducted on both synthetic and real images in both the grayscale and color domain prove the efficiency of the proposed model.

The rest of this paper is organized as follows:
Related studies are briefly described in section 2. In section 3, we detail the proposed energy functional of the proposed model. A comparison of the experimental results are then provided in section 4. Finally, some concluding remarks and a discussion of future studies are provided in section 5.
\label{sec:introduction}
\begin{figure}[h]
\centering
\includegraphics[scale=0.5]{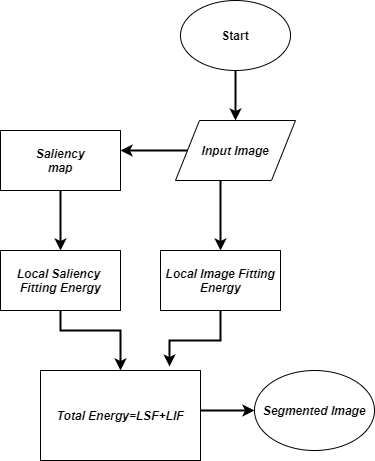}
\caption{Flow chart of proposed model}
\label{flow}
\end{figure}
\section{Related Studies}
There are different geometric active contour models based on a level set formulation. Here, we present some models that are closely related to the proposed model. First, we describe the CV model [12], which is a major region-based active contour model that is based on global image information. The CV model has certain limitations in the segmentation of images with intensity inhomogeneity, and to overcome such limitations, some local image information based models have been proposed. Among them, LBF [41] and LIF [29] are prominent models for the segmentation of images having intensity inhomogeneity. After a brief description of LBF and LIF, a saliency-based active contour model, SDREL [38], used in the proposed model is discussed.
\subsection{Chan and Vese Model} 
The CV model is one of the best region-based active contour models that is independent of object boundaries and can segment images having weak boundaries and a low gradient. It is a powerful and flexible method that can segment a variety of images, including some that are extremely difficult to segment in terms of ``classical'' segmentation methods, such as thresholding [7-9] or gradient based methods [14-18]. This model is based on the Mumford-Shah functional [26] for segmentation, and it is widely used in the medical imaging field, particularly for the segmentation of the brain, heart, and trachea. The model is based on an energy minimization problem, which can be reformulated in the level set formulation, leading to an easier way to solve the problem. The original energy equation is transformed into a level set formulation.
The CV model has the following level set energy function:
\begin{multline}
\label{CV}
F_{CV} (C,c_1,c_2)=\lambda_1\int_{outside (C)}|I(x)-c_1|^2H_\epsilon(\phi(x))dx\\
+\lambda_2\int_{inside (C)}|I(x)-c_2|^2(1-H_\epsilon(\phi(x)))dx\\
+\mu\int_{\Omega}|\nabla H_\epsilon(\phi(x))|^2dx+v\int_{\Omega}H_\epsilon(\phi(x))dx,
\end{multline}

Here, the first two terms are external energies that affect the motion of the curve, whereas the remaining terms are internal energies used to smooth the evolving curve. As the remaining equation parameters, $outside (C)$ and $inside (C)$ represent the outside and inside regions of contour C, respectively; $c_1$ and $c_2$ are the mean intensities of the inner and outer regions, respectively; and $\lambda_1$, $\lambda_2$, and $\mu$ are three constant parameters with values $>0$. Similarly, $\epsilon$ shows a correspondent epsilon, and $H_\epsilon$ indicates the Heaviside function, which is defined as follows:
\begin{equation}
H_\epsilon(\phi(x))=\dfrac{1}{2}\left(1+\dfrac{2}{\pi}arctan\left(\dfrac{\phi}{\epsilon}\right)\right).
\label{heaviside}
\end{equation}

Similarly, to find the mean intensities, $c_1$ and $c_2$, the following equations are defined. 

\begin{equation}
c_1=\dfrac{\int_{\Omega}I(x)H_\epsilon(\phi(x))dx}{\int_{\Omega}H_\epsilon(\phi(x))},
c_2=\dfrac{\int_{\Omega}I(x)(1-H_\epsilon(\phi(x)))dx}{\int_{\Omega}(1-H_\epsilon(\phi(x)))}.
\end{equation}

Using the calculus of variations [42], the final evolving equation of the CV model is as follows:

\begin{multline}
\dfrac{\partial\phi}{\partial t}=-\lambda_1\delta_{\epsilon}(\phi)(I-c_1)^2+\lambda_2\delta_{\epsilon}(\phi)(I-c_2)^2\\
+\mu \delta_{\epsilon}(\phi) div\left(\dfrac{\nabla\phi}{|\nabla\phi|}\right)-v\delta_{\epsilon}(\phi),
\end{multline}

where the derivative of $H_\epsilon$ is $\delta_\epsilon(x)$, which is called the Dirac delta function and is defined as follows:

\begin{equation}
\delta_{\epsilon}(\phi)=\dfrac{\epsilon}{\pi(\phi^2+\epsilon^2)}.
\label{dirac}
\end{equation}

The input image is considered to be homogeneous in this model, and hence the model achieves the best results for homogeneous images. If an inhomogeneous image is provided as an input, the image segmentation performance of the model is significantly reduced. 
\subsection{Local Binary Fitted Model (LBF)} 
To overcome the limitations of the CV model, Li et al. [41] presented a region-based model that uses the local image information to segment images in the presence of intensity inhomogeneity. The major contribution of this study is the introduction of a local binary fitting energy with a kernel function, which enables the extraction of accurate local image information. The basic idea is to introduce a kernel function to define a local binary fitting energy in a variational formulation, and thus the local intensity information can be embedded into a region-based active contour model. The LBF energy functional is further incorporated into a variational level set formulation without the re-initialization proposed by Li et al. [27]. Therefore, re-initialization is not necessary in the proposed method.
Li et al. define the following kernel function:
\begin{equation}
K_\sigma(x-y)=\dfrac{1}{(2\pi)^{\dfrac{n}{2}}\sigma^n}\exp^{-\dfrac{|x-y|^2}{2\sigma^2}}
\end{equation}

This is often called a Gaussian Kernel or window function with standard deviation $\sigma$ used to control the localization property. 
The complete LBF energy function is described as follows:

\begin{multline}
F_{LBF}(C,f_1,f_2)\\=\lambda_1\int_{\Omega}K_{\sigma} (x-y)|I(y)-f_1(x)|^2H_\epsilon(\phi(y))dy\\
+\lambda_2\int_{\Omega}K_{\sigma} (x-y)|I(y)-f_2(x)|^2(1-H_\epsilon(\phi(y)))dy\\
+\mu\int_\Omega\dfrac{1}{2}(\nabla\phi(x)-1)^2dx+v\int_\Omega\delta_\epsilon(\phi (x) |\nabla\phi(x)|dx.
\end{multline}

Here, the first two terms represent the curve motion, and the remaining two terms are used to remove the need of re-initialization and to regularize the zero-level contour, respectively. Other parameters of the equation are as follows: $\lambda_1$, $\lambda_2$, and $\mu$ are constant coefficients with values $>0$. In addition, $H_\epsilon(\phi)$ and $\delta_\epsilon(\phi (x))$ are defined as the Heaviside function \eqref{heaviside} and Dirac delta function \eqref{dirac}, respectively.

\begin{equation}
f_1(x)=\dfrac{K_\sigma*\left[H_\epsilon(\phi)I(x)\right]}{K_\sigma*H_\epsilon(\phi)}
\end{equation}
and
\begin{equation}
f_2(x)=\dfrac{K_\sigma*\left[(1-H_\epsilon(\phi))I(x)\right]}{K_\sigma*(1-H_\epsilon(\phi))}
\end{equation}

Here, $f 1 (x )$ and $ f_ 2 (x )$ functions show the approximate local intensities of the image inside and outside of the contour, respectively. The final evolving equation of the LBF is as follows:

\begin{multline}
\dfrac{\partial\phi}{\partial t}=-\lambda_1\delta_{\epsilon}(\phi)\int_{\Omega}K_{\sigma}(x-y)|I(x)-f_1(y)|^2dx\\ 
+\lambda_2\delta_{\epsilon}(\phi)\int_{\Omega}K_{\sigma}(x-y)|I(x)-f_2(y)|^2dx\\
+v\delta_{\epsilon}div\left(\dfrac{\nabla\phi}{|\nabla\phi|}\right)+\mu\left(\nabla\phi-div\left(\dfrac{\nabla\phi}{|\nabla\phi|}\right)\right),
\end{multline}

This model provides prominent results in terms of image segmentation with intensity inhomogeneity. However, four convolution operations in every iteration increase the computational cost. Zhang et al. [29] overcame the drawbacks of LBF by presenting a model that provides results equivalent to those of LBF but with fewer computations.
\begin{figure}[]
 \includegraphics[scale=0.5]{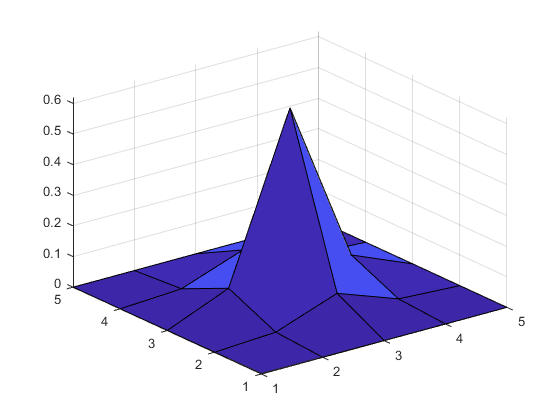}
 \caption{3D plot of Gaussian kernel with standard deviation 0.5 and kernel's size 5.}
 \label{gaussian}
 \end{figure}
\begin{figure*}
 \centering
 \includegraphics[scale=0.22]{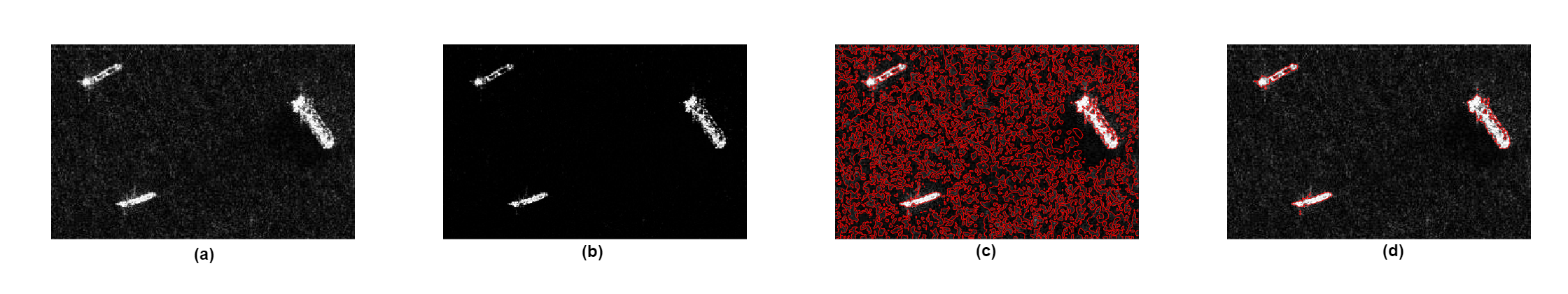}
 \caption{Image segmentation with and without a saliency map in an active contour model. The first image is an original image, and the second is its saliency map. The last two images are the results without and with saliency, respectively.}
 \label{saliency_effect}
 \end{figure*}
\begin{figure}[]
 \includegraphics[scale=0.22]{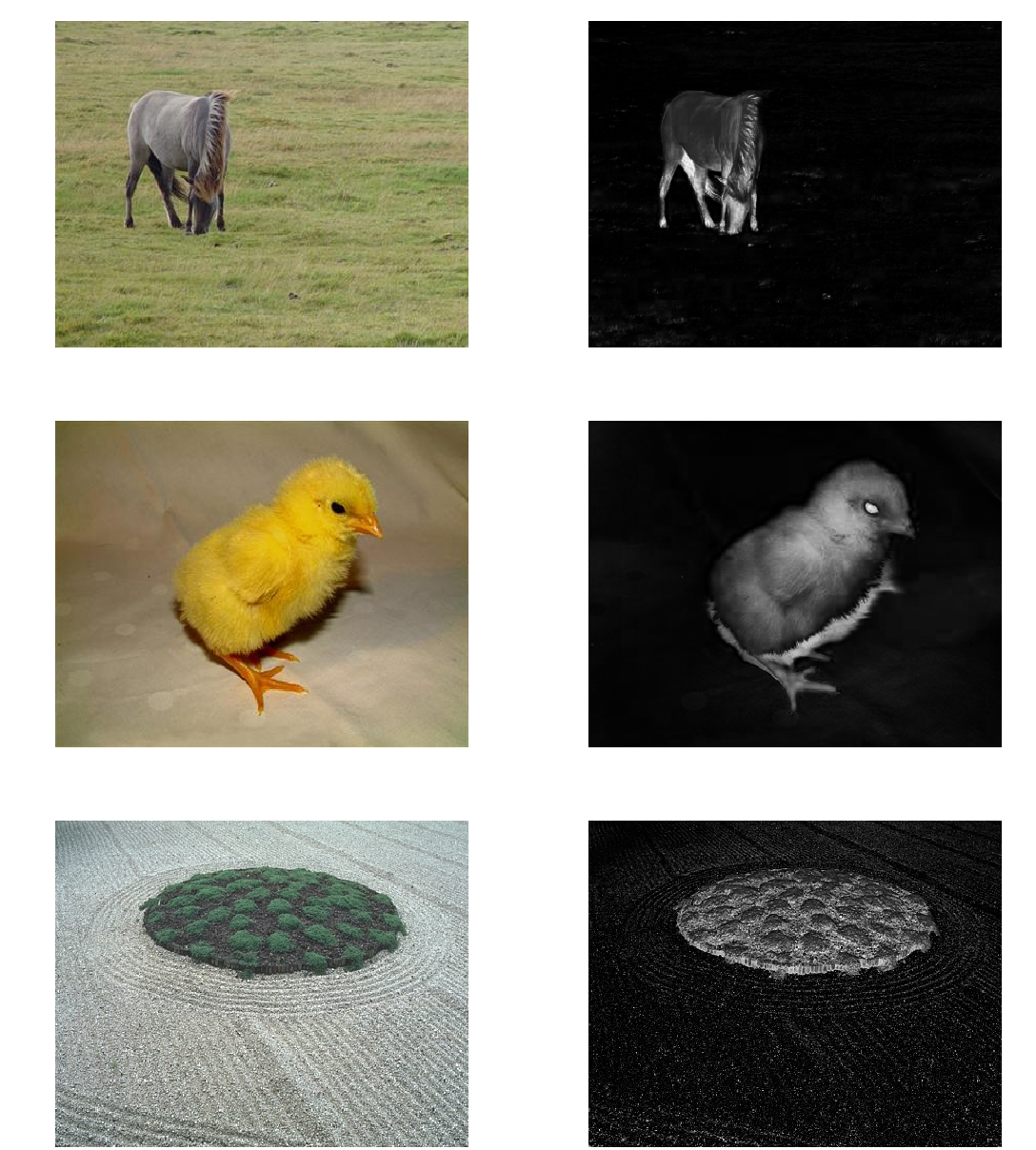}
 \caption{The first column shows the input images, and the second column shows their saliency map.}
 \label{saliency}
 \end{figure}
\begin{figure*}
\centering
\includegraphics[scale=0.2]{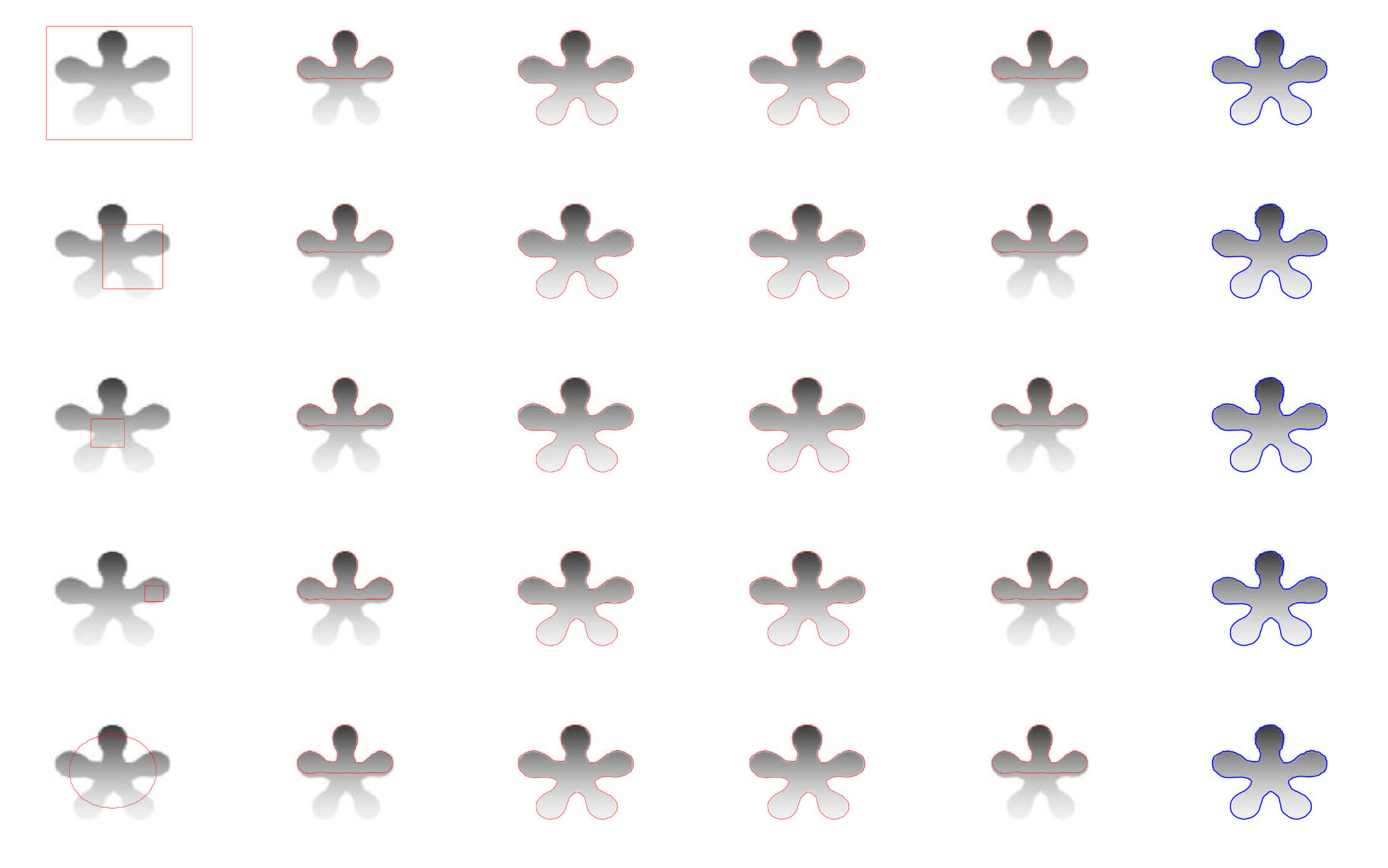}
\caption{Comparison of segmentation results of the proposed and other models on a synthetic image. From left to right, the columns represent original images using a different initial contour and the results of the CV, LBF, LIF, SDREL, and proposed models.}
\label{star}
\end{figure*} 
\begin{figure}
\includegraphics[scale=0.25]{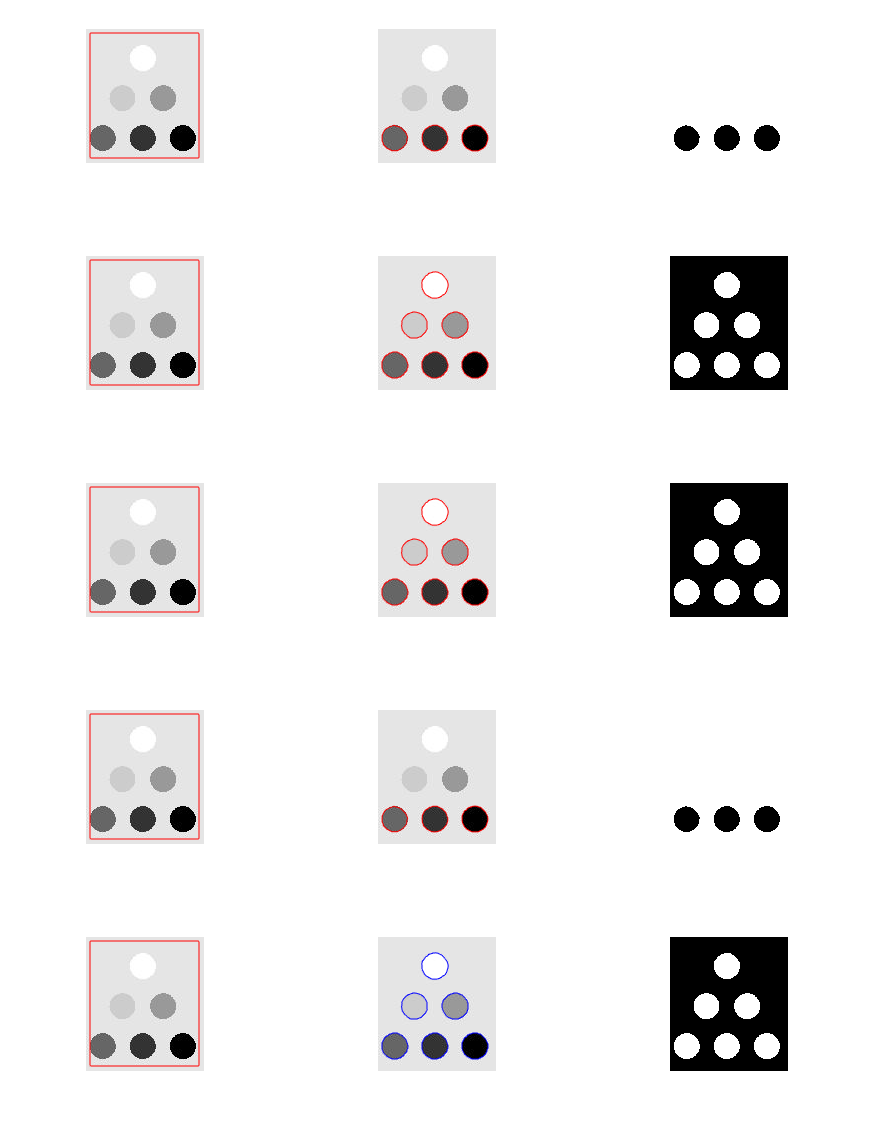}
\caption{Segmentation of six objects in one image each with a different intensity. The first column is the original image with the initial red contour. The second column shows the results of the CV, LBF, LIF, SDREL, and proposed models from top to bottom. The last column shows the binary result of column two.}
\label{hole}
\end{figure}
\begin{figure}
\includegraphics[scale=0.25]{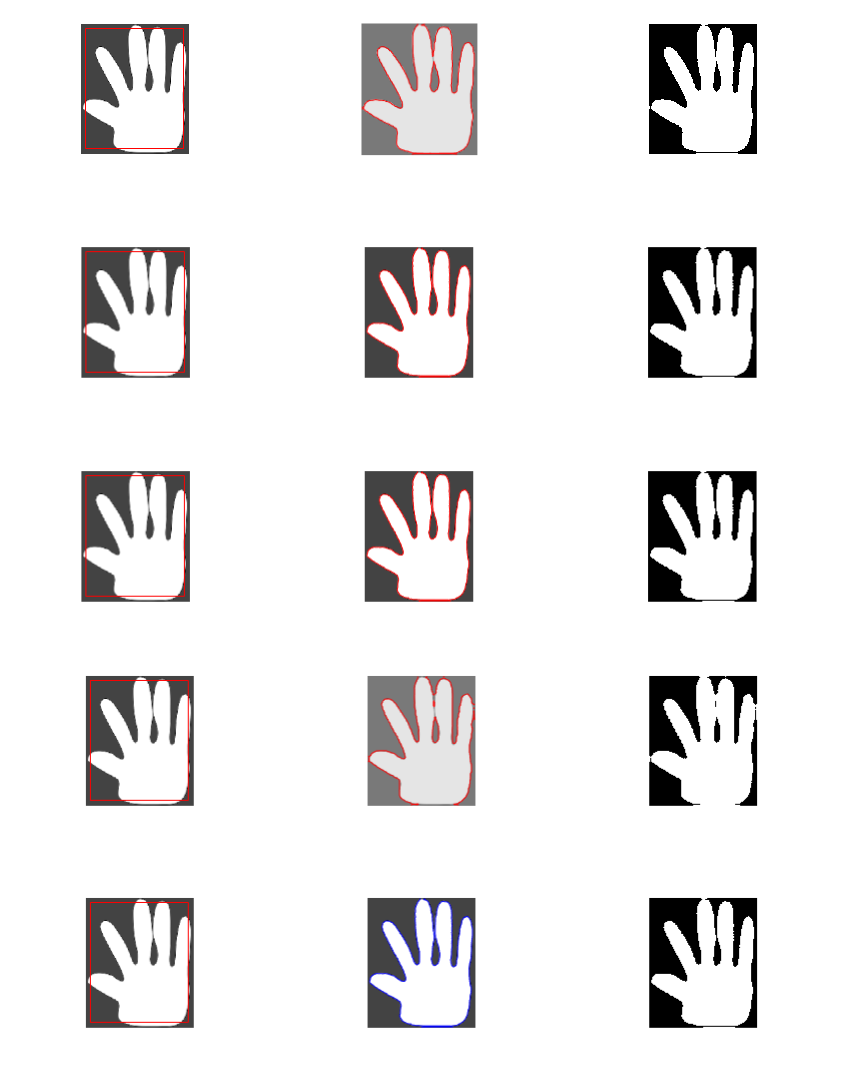}
\caption{Comparison of the segmentation results of the proposed and other models on a complex image with a constant intensity. The first column is the original image with the initial red contour. The second column shows the results of the CV, LBF, LIF, SDREL, and proposed models from top to bottom.}
\label{hand}
\end{figure}
\begin{figure*}[h]
\centering
\includegraphics[scale=0.23]{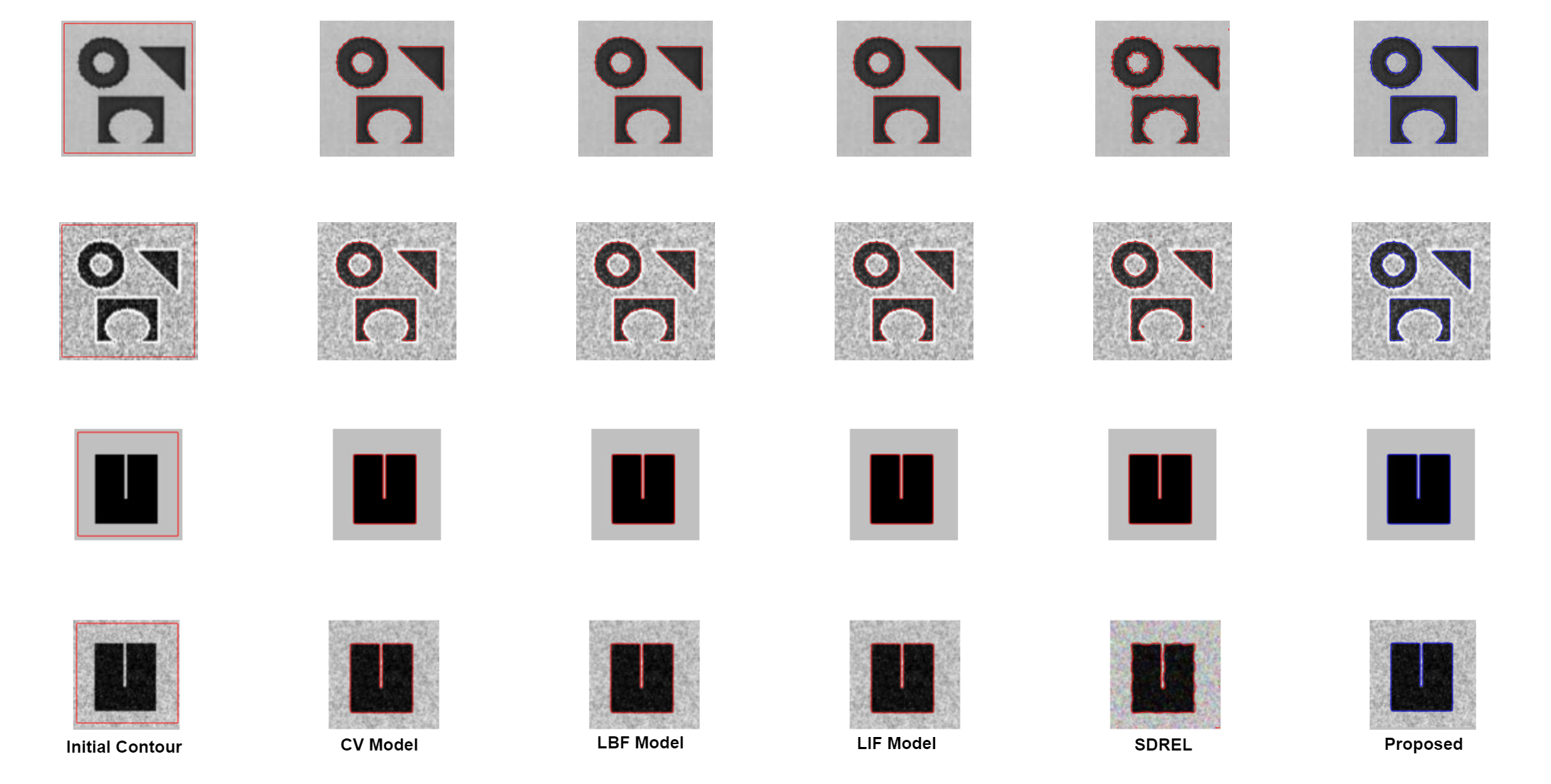}
\caption{The results on synthetic images with and without noise. The first and third rows show the results on a synthetic image without noise, whereas the second and fourth rows are the results with noise.}
\label{noisy}
\end{figure*}
 
\subsection{Local Image Fitting Model (LIF)} 
Zhang et al. [29] provided two main contributions to the research on active contours. First, they presented an LIF model that is also based on local image information as LBF but with less computational cost. Second, they introduced a Gaussian kernel that is used to smooth the level set function and remove the need for re-initialization. The main idea of this model is to reduce the difference between a local fitted image and the original image. Mathematically, an LIF model can be represented as follows:
\begin{equation}
E_{LIF}=\dfrac{1}{2}\int_\Omega|I(x)-I^{LFI}(x)|^2dx.
\end{equation}
In addition, the local fitted image is calculated in the following manner:
\begin{equation}
\label{LFI}
I^{LFI}(x)=m_1(x)H_\epsilon(\phi)+m_2(x)(1-H_\epsilon(\phi)),
\end{equation}
where $m_1$ and $m_2$ are local mean intensities and are computed as follows:
 \begin{equation}
\begin{aligned}
\label{m1,m2}
&m_1=mean(I \in ( ( x \in \Omega \mid \phi(x) < 0  ) \cap  Wk(x)))\\&
m_2=mean(I \in (( x \in \Omega \mid  \phi(x)> 0) \cap Wk(x))).
\end{aligned}
\end{equation}

Here, $Wk(x)$ is a truncated Gaussian window with size $(4k+1)×(4k+1)$, and its deviation is $\sigma_m$. In addition, $k$ is the largest integer smaller than $\sigma_m$. Using the calculus of variations, the final evolving level equation is as follows:

\begin{equation}
\label{LIF}
\dfrac{\partial\phi}{\partial t}=\left(I(x)-I_{LFI}(x)\right)\left(m_1(x)+m_2(x)\right)\delta_{\epsilon}(\phi),
\end{equation}

Here, $\delta_{\epsilon}(\phi$ represents a Dirac function defined in \eqref{dirac}. The segmentation result of the LIF model is as good as that of the LBF model but faces a certain limitation, for example, the sensitivity of the initial contour position and a stop of the contour at the local minima. 
\subsection{SDREL}
Visual saliency is the perceptual quality that makes an object, person, or pixel stand out relative to its neighbors, and thus captures our attention. A saliency representation of an image with an active contour model shows prominent results on the grayscale and color images. There are different methods[43-45] to find the saliency map of an image, each of which has its own advantages and disadvantages. SDREL [38] is a saliency driven region based active contour model that merges a saliency map with an active contour to segment the targeted objects.

In this model, a saliency map of the original image is found using the approach in [40], and then it is applied with the CV model to compute the final energy functional. An edge term is also used to enhance the segmentation results. The level set equation of SDREL is as follows:
\begin{multline}
\dfrac{\partial\phi}{\partial t}=g[-\lambda_1.(I-c_1)^2+\lambda_2(I-c_2)^2\\
\alpha_1(S-s_1)^2+\alpha_2(S-s_2)^2].
\end{multline}

Here, the first two terms are the original CV energy equation \eqref{CV}, and the remaining two are a saliency driven CV model. In addition; $\lambda_1$, $\lambda_2$, $\alpha_1$, and $\alpha_2$ are constant positive parameters; $g$ is an edge term used to promote the boundaries of weak objects; $S$ is the saliency map of the original image; and $s_1$ and $s_2$ are the mean saliency of the inside and outside areas of the contour, respectively.

The edge term of SDREL is calculated using the following equation:
\begin{equation}
g=\dfrac{1}{1+|\nabla G_\sigma *I|^2}
\end{equation}

Here, I represents the input image, $\nabla$ represents the gradient operator, and $G_\sigma$ indicates a Gaussian kernel with a standard deviation of $\sigma$, the default value of which is set to 0.8 and the size of the kernel is set to 3.\\Similarly, saliency map $S$ and the mean $s_1$ and $s_2$ are computed by the following:
\begin{equation}
 S(x,y)= |I_u-I_g(x,y)|
\end{equation}

Here, $I_u$ is the mean of the input image and $I_g$ is a Gaussian blurred image.

\begin{equation}
s_1=\dfrac{\int_{\Omega}S(x)H_\phi(x))dx}{\int_{\Omega}H_\phi(x))},
s_2=\dfrac{\int_{\Omega}S(x)(1-H_ \phi(x)))dx}{\int_{\Omega}(1-H_\phi(x)))}.
\end{equation}

The SDREL model uses the saliency map in terms of the active contour to segment the images and obtain prominent results for both grayscale and color images. The SDREL model uses only the global region information of an image, such as the intensity of an image and its saliency map and does not consider local region information. Thus, owing to the global behavior of the active contour model, it is unable to extract an object with intensity inhomogeneity and faces the same limitations as the CV model.
\section{Proposed Method} 
In this paper, the active contour driven by the LSF and LIF energies is proposed for the image segmentation of homogeneous and inhomogeneous images. First, a saliency map of an input image is calculated. Then, LSF
 energy is derived using saliency map, and at the end final novel energy functional is proposed that uses LIF energy with LSF energy. Finally, the variational and gradient descent flow methods are adopted to minimize the total energy functional for segmentation. A flow chart of the proposed model is illustrated in Fig. \eqref{flow}.
\subsection{Saliency Map}
As discussed above, there are many methods to compute the saliency map of an input image; however, in this study, we compute the saliency map using the approach in [40]. Fig. \eqref{saliency} shows the saliency map of different images. These images are randomly taken from MSRA dataset[47].
\begin{equation}
\label{s1}
 S(x,y)= |I_u-I_g(x,y)|
\end{equation}  

Here, $I_u$ is called the arithmetic mean of the input image, and $I_g(x,y)$ is a blurred version of the input image that uses a Gaussian filter to eliminate noise and disturbances in corrupted images. In this case, a Gaussian filter is defined as follows:
 \begin{equation}
 \label{s2}
 I_g(x,y)= G_\sigma *I(x,y)
\end{equation}

Here, $G_\sigma$ indicates a Gaussian kernel with a standard deviation of $\sigma$. The standard value of $\sigma$ and the size of the kernel are 0.5 and 5, respectively. Fig. \eqref{gaussian} shows 3D plot of the Gaussian kernel. After computing the saliency map, we calculate the saliency driven LSF energy.
\begin{figure*}[h]
\centering
\includegraphics[scale=0.25]{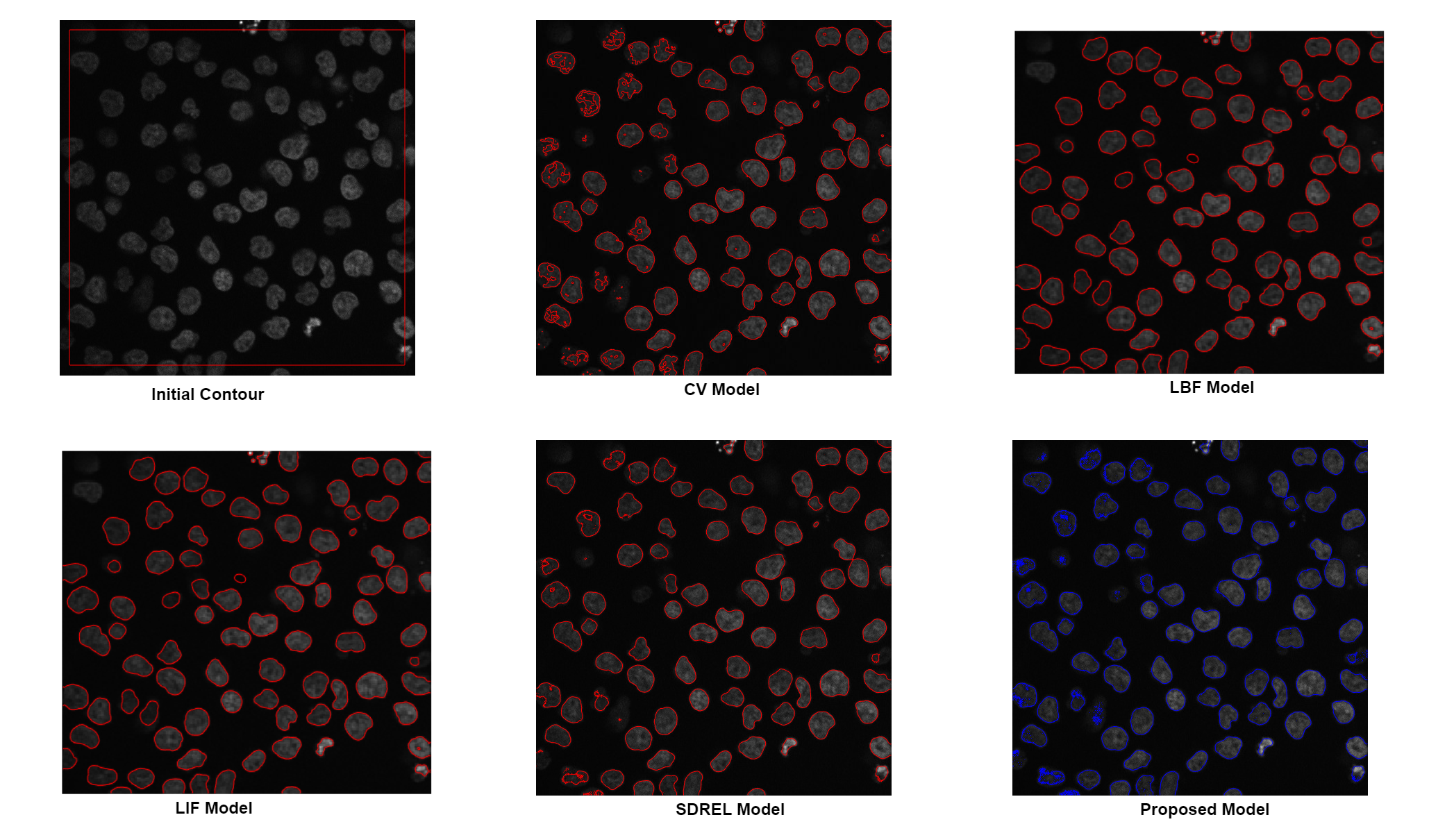}
\caption{Comparison of segmentation results of microscopic cells.}
\label{cell}
\end{figure*}
\begin{figure*}[h]
\centering
\includegraphics[scale=0.15]{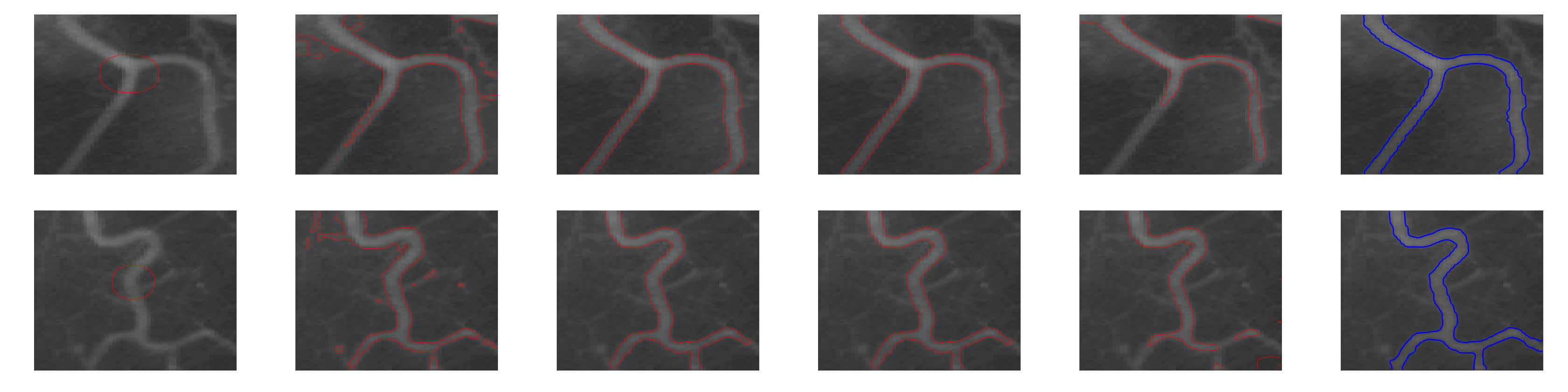}
\caption{Segmentation results of two real blood vessel images. The first column shows two blood vessels with the initial contour, and the second to last columns are the results of the CV, LBF, LIF, SDREL, and proposed models, respectively.}
\label{vessel}
\end{figure*}
\begin{figure*}[h]
\centering
\includegraphics[scale=0.15]{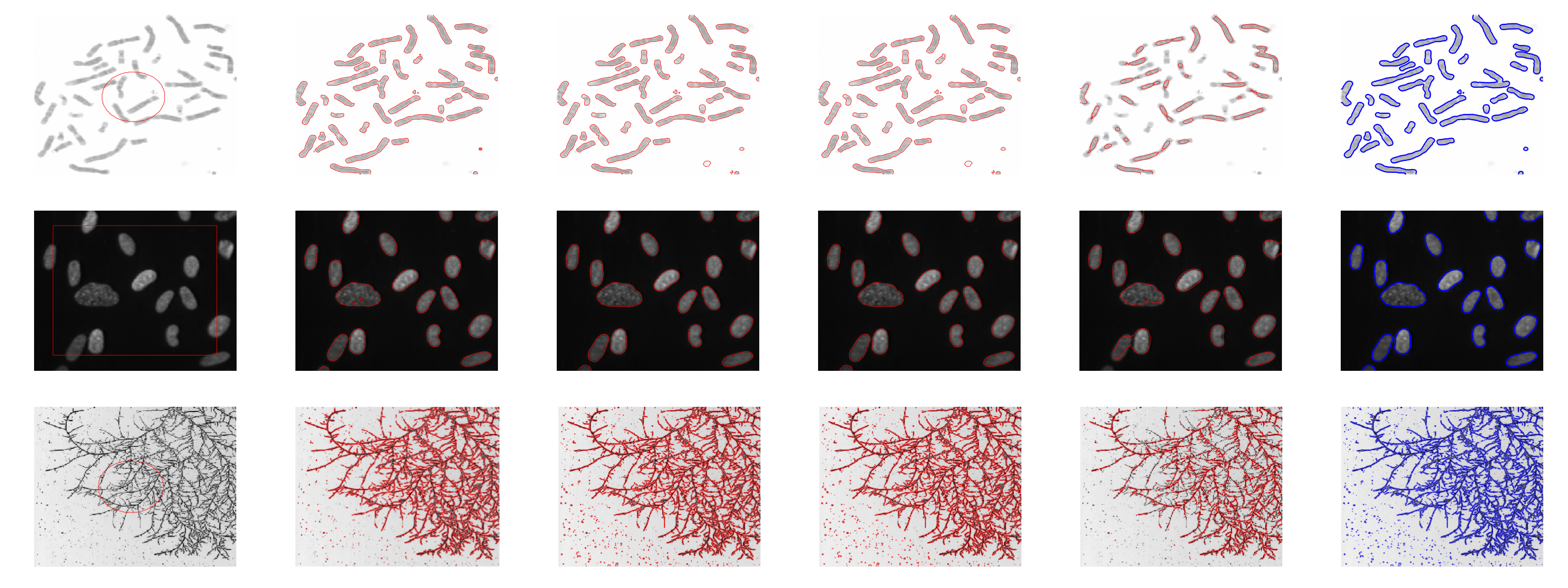}
\caption{Segmentation results of bio-medical grayscale images. The first column shows three randomly chosen images with the initial contour, and the second to the last columns represent the results of the CV, LBF, LIF, SDREL, and proposed models, respectively.}
\label{fungus}
\end{figure*}
\subsection{Local Saliency Fitting Energy}
It is distinctly understood that the use of only the global-saliency based energy [38] is insufficient to segment objects when the foreground and background regions of the images are inhomogeneous. To obtain a better level set based image segmentation model, inspired by the advantages and disadvantages of the above-mentioned level set models, we present a novel saliency-based local image energy called LSF energy, which is computed by embedding saliency into the LIF model. It was experimentally proven that prominent results are achieved not only for grayscale images but also for color images having intensity inhomogeneity.
Therefore, using the advantages of an LIF and a saliency map, we propose the following energy.
\begin{equation}
\begin{aligned}
\label{ELSF1}
E_{LSF}=\dfrac{1}{2}\int_\Omega|S(x)-S_{LSF}(x)|^2dx,
\end{aligned}
\end{equation}

Here $S(x)$ is a saliency map of the image computed by \eqref{s1} and $S_{LSF}$ is a saliency driven local fitted image, which is defined as follows:
\begin{equation}
\begin{aligned}
\label{LSF}
S_{LSF}(x)=L_{s1}(x)H_1+L_{s2}(x)H_2.
\end{aligned}
\end{equation}

Here, $H_1=H_\epsilon(\phi)$, $H_2=1-H_\epsilon(\phi)$, and $H_\epsilon(\phi)$ are the Heaviside functions defined in \eqref{heaviside}, and $L_{s1}$ and $L_{s2}$ are the saliency-based intensity mean of the inside and outside areas of the region, respectively.
The values of $L_{s1}$ and $L_{s2}$ are defined as follows:
\begin{equation}
\begin{aligned}
\label{s1,s2}
&L_{s1}=mean(S \in ((x \in \Omega \mid \phi(x)< 0) \cap  Wk(x)))\\&
L_{s2}=mean(S \in ((x \in \Omega \mid  \phi(x)> 0) \cap Wk(x))).
\end{aligned}
\end{equation} 

where $Wk(x)$ is the truncated Gaussian window with size $(4k+1)×(4k+1)$, and its deviation is $\sigma_s$. In addition, $k$ is the biggest integer smaller than $\sigma_s$.
Using the calculus of variations, we minimize $E_{LSF}$ (\ref{ELSF1}) with respect to $\phi$ to obtain the following equation:
\begin{equation}
\begin{aligned}
\label{ELSF}
\dfrac{\partial\phi}{\partial t}=(S(x)-S_{LSF}(x))(L_{s1}(x)-L_{s2}(x))\delta_\epsilon(\phi),
\end{aligned}
\end{equation}
where $\delta_\epsilon(\phi)$ is the regularized Dirac delta function defined in \eqref{dirac}.   
\subsection{Final Level Set Equation}
The final level set model is composed of the LSF energy $E_{LSF}$ \eqref{ELSF} and LIF energy $E_{LIF}$ \eqref{LIF}.
\begin{equation}
\begin{aligned}
\label{final1}
E(\phi)=E_{LSF}(\phi)+E_{LIF}(\phi)
\end{aligned}
\end{equation}

To ensure a stable evolution of the level set function $\phi$, we add the distance regularizing term to penalize the deviation of the level set function $\phi$ from a signed distance function. The deviation of the level set function $\phi$ from a signed distance function is characterized by the following integral:
\begin{equation}
\begin{aligned}
\label{p}
P(\phi)=\int_\Omega\frac{1}{2}(|\nabla\phi(x)|-1)^2dx
\end{aligned}
\end{equation}

To regularize the zero-level contour of $\phi$, we also need the length of the zero-level curve (surface) of $\phi$, which is given by the following:
\begin{equation}
\begin{aligned}
\label{l}
L(\phi)=\int_\Omega\delta\phi(x)|\nabla\phi(x)|dx
\end{aligned}
\end{equation}

By minimizing equations $\ref{p}$ and $\ref{l}$ with respect to $\phi$, similar to Li et al.'s approach [28], we finally obtain the final variational formulation as follows:
\begin{multline}
\label{final2}
\dfrac{\partial\phi}{\partial t}=[S(x)-S_{LSF}(x))(L_{s1}(x)-L_{s2}(x))\delta_\epsilon(\phi)]\\+[(I(x)-I_{LIF}(x))(m_1(x)-m_2(x)\delta_\epsilon(\phi)]\\+v\delta_{\epsilon}div\left(\dfrac{\nabla\phi}{|\nabla\phi|}\right)+\mu\left(\nabla\phi-div\left(\dfrac{\nabla\phi}{|\nabla\phi|}\right)\right),
\end{multline}

Here, the first two terms are the saliency driven and local image energies used to move the contour toward the boundaries of the object, whereas the third term is the length of the contour, and the last term represents the penalizing term used to remove the need for contour re-initialization. In addition, $\mu$ and $\nu$ are positive parameters applied to adjust the penalizing and length terms, respectively.

\subsection{For Color Images}
The saliency driven active contour level set energy functional can easily segment color images. To extend the original energy functional (\ref{final2}) for color images, we compute the saliency map as follows:
\begin{equation}
 \label{scolor}
 S(x,y)= \parallel I_u-I_g(x,y)\parallel
\end{equation}

where $I_{u}$ is the mean image feature vector, $I_{g}$ is a Gaussian blurred image pixel vector, and $\parallel\bullet\parallel$ is the L2 norm. Using the Lab color space, each pixel location is an $[L;a;b]^T$ vector, and the L2 norm is the Euclidean distance. The values of ($S-S_{LSF}$), ($L_{s1}-L_{s2}$), $(I-I_{LIF})$, and $(m_{1}-m_{2})$ are replaced with the Euclidean vector norms in the original energy equation \eqref{final2}. The minimization of the grayscale image equation \eqref{final2} is the same.
 \begin{figure*}[]
 \centering
 \includegraphics[scale=0.15]{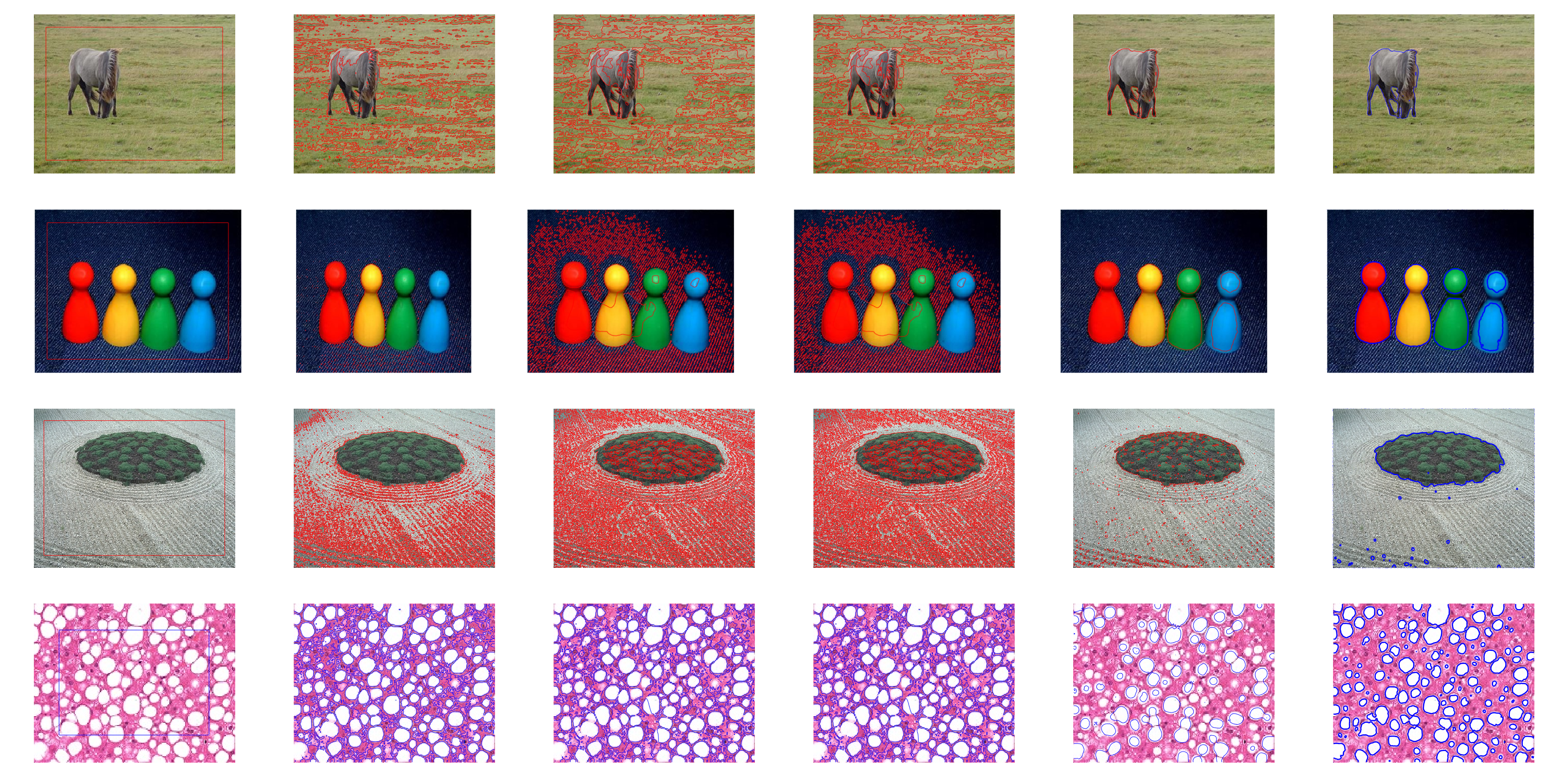}
 \caption{Comparison of segmentation results of color images. The first column shows an image with the initial contour and the second to the last columns represent the results of the CV, LBF, LIF, SDREL, and proposed models, respectively.}
 \label{bsd500}
 \end{figure*}
  \begin{figure*}[]
 \centering
 \includegraphics[scale=0.2]{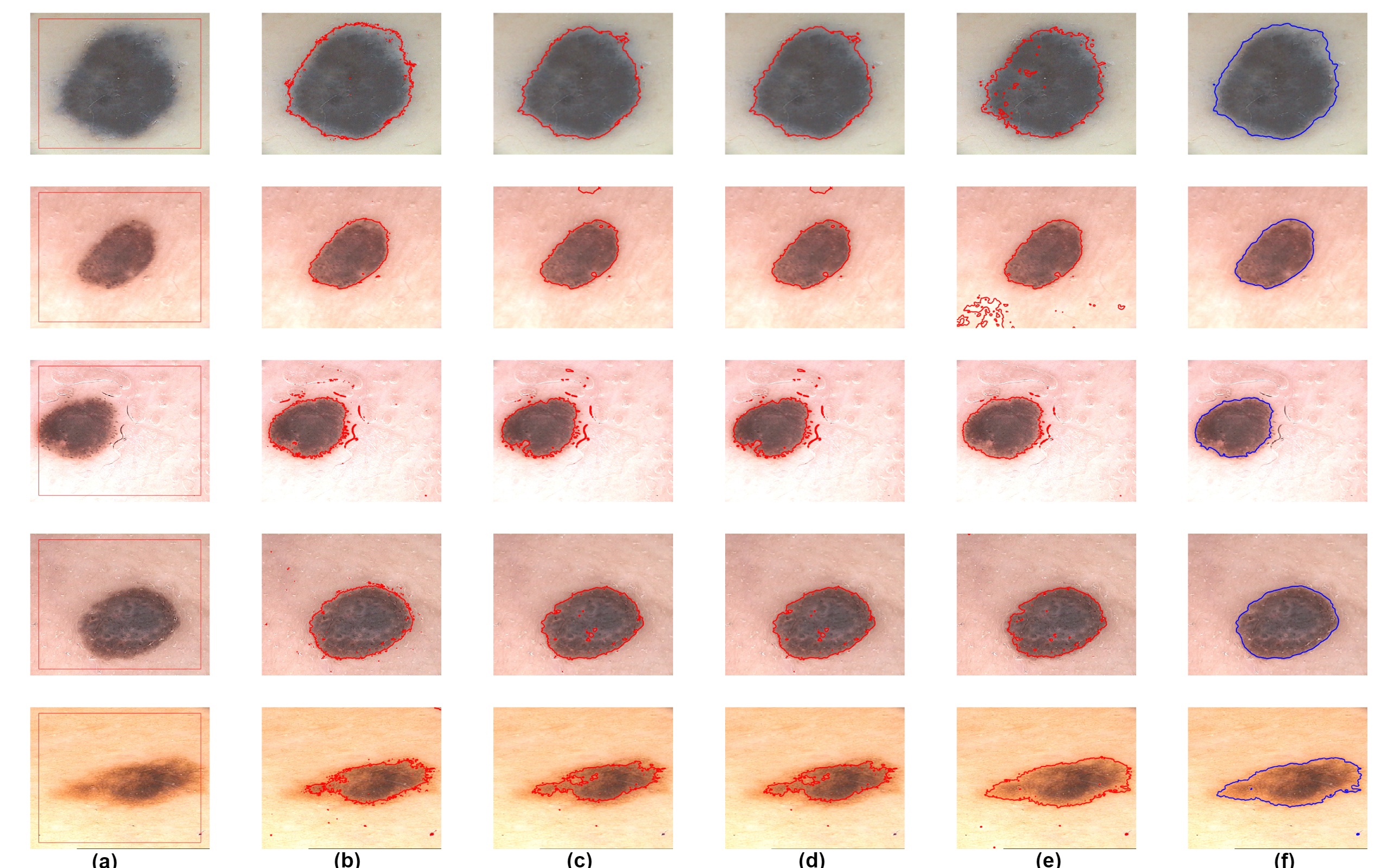}
 \caption{Experiment results of dermoscopic images. The first column shows the original image with the initial contour, and the second to the last columns represent the results of the CV, LBF, LIF, SDREL, and proposed models, respectively.}
 \label{dermoscopic}
 \end{figure*}
\subsection{Contributions}
The experiment using the proposed model demonstrates that it yields far more prominent results on all types of images with different color intensities. The main contributions of this model are as follows:
\begin{itemize}
	\item This model provides a unique energy functional that is driven by the saliency map of an image and the image itself.
	\item A novel saliency driven local saliency fitting energy term is proposed. 
	\item Using the advantages of a visual saliency map and a local image fitting model, we designed a new saliency-based level set energy functional that has sufficient strength to improve the level set based image segmentation.
	\item The regularization and penalizing terms are used for a smooth contour movement and to remove the need for contour re-initialization, respectively.
	\item The new model is applicable to homogeneous and inhomogeneous grayscale images and color images.
\end{itemize}
 The near-optimal parameter settings derived are represented in Table \eqref{parameter}. The algorithm for the implementation of proposed model is described in Algorithm \ref{Alg1}.
\begin{table}[h]
\caption{Parameters}
\begin{tabular}{lll}
\hline
\multicolumn{1}{|l|}{Parameter}   & \multicolumn{1}{l|}{Name}                   & \multicolumn{1}{l|}{Value}             \\ \hline
\multicolumn{1}{|l|}{$\Delta{t}$} & \multicolumn{1}{l|}{Time step}              & \multicolumn{1}{l|}{$0.1$}            \\
\multicolumn{1}{|l|}{$p$}         & \multicolumn{1}{l|}{Initial level set}      & \multicolumn{1}{l|}{$2$}               \\
\multicolumn{1}{|l|}{$\sigma_s$}    & \multicolumn{1}{l|}{Standard Deviation}        &
 \multicolumn{1}{l|}{$3.5$}             \\
 \multicolumn{1}{|l|}{$\sigma_m$}    & \multicolumn{1}{l|}{Standard Deviation}        &
 \multicolumn{1}{l|}{$3$}             \\
\multicolumn{1}{|l|}{$\mu$}         & \multicolumn{1}{l|}{penalizing term} & 
\multicolumn{1}{l|}{$1$} \\
\multicolumn{1}{|l|}{$\epsilon$}    & \multicolumn{1}{l|}{Epsilon}        & \multicolumn{1}{l|}{$1.5$}             \\
\multicolumn{1}{|l|}{$\nu$}         & \multicolumn{1}{l|}{Length term} & 
\multicolumn{1}{l|}{$0.001*255*255$} \\
\hline                            
\end{tabular}
\label{parameter}
\end{table} 
\begin{algorithm}
\caption{Proposed Model} \label{Alg1}
\begin{algorithmic}[1]
\STATE Initialize the level set function $\phi$ and set the constant functions as the initial contour.
\begin{equation}
\phi_{t=0}=\begin{cases}
-p, & x\in\Omega_0-\partial\Omega_0\\
 0, &x \in\partial\Omega_0\\
 p, & x\in\Omega-\Omega_0.
  \end{cases}
\end{equation}
where $p$ is a constant parameter, $\Omega_0$ is a subset of the image domain $\Omega$, and $\partial\Omega_0$ is the boundary of $\Omega_0$. 
\STATE Compute the saliency map of the input image using \eqref{s1}.
\STATE Compute $H_{\epsilon}(\phi)$ and $\delta_{\epsilon}(\phi)$ using Equations \eqref{heaviside} and \eqref{dirac}, respectively.
\STATE Find the image means $m_1$ and $m_2$ and the saliency means $L_{s1}$ and $L_{s2}$ using \eqref{m1,m2} and \eqref{s1,s2}.
\STATE Calculate $S_{LSF}$ and $I_{LIF}$ using Equations \eqref{LSF} and \eqref{LFI}, respectively.
\STATE Calculate $E_{LSF}$ and $E_{LIF}$ using Equations \eqref{ELSF} and \eqref{LIF}, respectively.
\STATE Compute the penalizing term using \eqref{p} and the length term using \eqref{l}, to smooth the level set function and remove the need for re-initialization.
\STATE Compute the final level set evolution using Equation \eqref{final2}.
\STATE Repeat the steps 3-8 till segmentation. 
\end{algorithmic}
\end{algorithm}
\section{Experiment Results}
The proposed model was tested on homogeneous and in-homogeneous grayscale and color images. We tested the model on synthetic images, real images, and different datasets. We compared the results with related active contour models (CV, LBF, LIF, and SDREL) and their implementations are publicly available at [49-52]. The experiment was carried out on MATLAB 2020(a) installed on a PC running 64-bit OS (Windows 10) with a 3.00 GHz Intel Core i5-8500 CPU and 8 GB RAM. The same system was used to implement the proposed model and compare its efficiency against other models. We followed the previously defined algorithm using the parameters in Table \eqref{parameter}.
\subsection{Effect of Saliency Map}
Using a saliency driven active contour model, we expected to achieve a more reliable and useful segmentation. Fig. \ref{saliency_effect}(a) shows the original sub-image of TerraSAR-X data with three ships. The segmentation results show a comparison of the active contour model with and without a saliency term. It can be clearly observed from figure \eqref{saliency_effect} that the active contour model without a saliency map cannot segment the ships from the background, and a disturbance occurred owing to the image inhomogeneity and noise recognized as part of the foreground. By contrast, the result of the model with the saliency term was much more reliable and only ships targeted in the foreground of the image were recognized. 
\begin{figure*}[]
	\centering
	\includegraphics[scale=0.7]{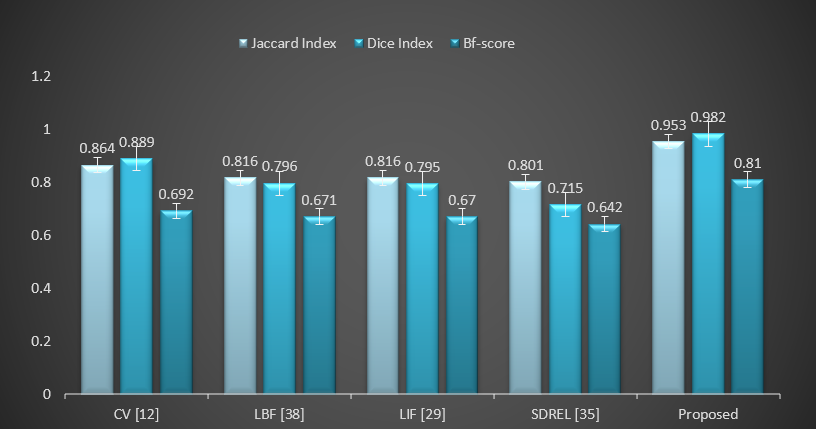}
	\caption{Jaccard index, Dice index, and BFscore of $PH^2$ images}
	\label{JI}
\end{figure*}
\subsection{Synthetic images}
To verify the validity of the proposed model, for a contour initialization, and to check the model at different image intensities, we conducted experiments on different synthetic images, as shown in Figs. \eqref{star}, \eqref{hole}, \eqref{hand}, and \eqref{noisy}. The first image in Fig. \eqref{star} was tested with five different contour positions. The first column of the figure represents an original image with distinct initial contours, whereas the second to the last columns represent the results of the CV, LBF, LIF, SDREL, and proposed models, respectively. The experimental results show that the proposed model is not dependent on the initial contour and is also compatible for images with intensity inhomogeneity. 

We then take two different synthetic images, an image with six different intensities, as shown in Fig. \eqref{hole}, and complex image of a hand with a constant intensity, as shown in Fig. \eqref{hand}. We also provide the binary results of final contour evolution to clarify the experiment. As shown in Figs. \eqref{hole} and \eqref{hand}, the first column is of the initial contour, the second is the contour fitted result of all models, and the third is the binary result of the contour fitted image. Fig. \eqref{hole} with six different intensities shows that the CV and SDREL methods, which depend on global image information, only segment an object having intensity homogeneity, whereas the proposed model and the LBF and LIF models segment all six objects with different intensities. The experimental results show that the proposed model is capable of segmenting images with both homogeneous and inhomogeneous intensities. Fig. \eqref{hand} shows an image of a hand with joined middle fingers. Here, all state-of-the-art models segment the hand from the background; however, the CV and SDREL methods are unable to segment the joined middle fingers precisely.

Furthermore, we tested our model on synthetic images with and without noise. Consider Fig. \eqref{noisy} in which the first row shows the results of all models on the original image without noise, the second row shows results with noise, the third row shows the original image, and the last row shows the result of an image with noise. The experimental results indicate that the proposed model has no effect on the amount of noise, and the segmentation of the object is the same for both types of images.

\subsection{Real Images}
This section describes the efficiency of the proposed model on some real-world images. For instance, consider a microscopic image of cells in grayscale, as shown in Fig. \eqref{cell}, from which we want to extract all the cells from the background using image segmentation tools. Furthermore, the cells have weak boundaries and a highly related intensity as the background. Therefore, it is difficult to extract these cells from a black background using an ordinary model. Here, we need a model capable of extracting all the microscopic cells from a dark background. The experiment results show that, while using a global region-based model for cell detection, such as the CV model, the segmentation is poor, whereas a local region based model such as an LBF or an LIF extract the cells more precisely. However, they also miss some cells that have almost the same intensity as the background. Although the SDREL model uses global image information, owing to its edge-based term it also segments many cells from the background. By contrast, results of the proposed model indicate that it segments almost all the microscope cells from the black background with more accuracy and speed. 

Similarly, Fig. \eqref{vessel} shows the segmentation results of two blood vessel images having intensity inhomogeneity and extremely weak boundaries. As global region based models, CV and SDREL are unable to segment the vessels from the background, whereas the local region based models such as the LBF and LIF models provide interesting results; however, the proposed model shows better results than all of the previous methods and segments the blood vessel with more accuracy and precision.

To further verify the efficiency of our model in terms of image segmentation, we conducted experiments on a bio-medical grayscale image dataset [46]. As shown in the first row of Fig. \eqref{fungus}, we attempted to extract the chromosome, whereas in second row, we tried to find algae from a dark background. Finally, the last row shows the segmentation of a fungus. These bio-medical grayscale images have different variations in intensity, which make it difficult to segment. From Fig. \eqref{fungus}, we can see that the proposed model conveniently extracts all the required objects from these bio-medical images.
\subsection{color images}
We further tested our model by segmenting color images. Figs. \eqref{bsd500} and \eqref{dermoscopic} show the segmentation of the color image datasets. In Fig. \eqref{bsd500}, the first three images are chosen from the MSRA 1000 dataset [44][47]. The first column shows the original image with the initial contour, whereas the remaining columns are the results of the CV, LBF ,LIF, SDREL, and proposed models. The experimental results show that the CV, LBF, and LIF models are unable to segment the object from the color images, whereas SDREL shows prominent results. However, the segmentation results of our model on the MSRA dataset are more accurate and reliable. Similarly, the last row of Fig.\eqref{bsd500} shows a fatty liver image in which the purpose of the segmentation is to extract the fatty circle from the image. All the applied models extract some parts of the fats from the liver, whereas our model achieves the best segmentation and extracts all fat circles from the liver image. However we still have gap to improve the segmentation accuracy of the region based ACMs such color images.

The increasing incidences of melanoma, a type of skin cancer, have recently promoted the development of computer-aided diagnosis (CAD) systems for the segmentation of dermoscopic images. Because $PH^2$ is a dermoscopic image database, we also tested our model on a dataset of dermoscopic images [48]. Fig. \eqref{dermoscopic} shows the experiments conducted to segment melanoma from the skin. Here, the first column shows the initial contour, and the second to the last columns are the results of the CV, LBF, LIF, SDREL, and proposed models, respectively. All previous model results on the $PH^2$ images are acceptable, although in some images these models show false contours and noise, whereas the proposed model clearly segments the melanoma from the skin.

\subsection{Evaluation Measurements}
Furthermore, the accuracy of the segmentation results can be assessed quantitatively using an image segmentation evaluation measurement. The most frequently used segmentation evaluation techniques are the Dice Index, Jaccard similarity index (Jaccard index), and contour matching score (bfscore). The same system and data are used to measure these evaluations for all models discussed above. 

The Dice Index, often referred as an overlap index, is calculated by overlapping the segmentation mask and the ground truth mask. The value of the Dice Index indicates by how much the segmentation result is equivalent to the actual result. The range of the Dice Index is between [0, 1], where a value closer to 1 shows a more accurate segmentation. The dice Index is calculated using the segmentation result (SR) and ground truth (GT). The formula used to calculate the Dice Index is as follows:
\begin{equation}
Dice(SR,GT)= \frac{2|S_r \cap S_g| }{|S_g| + |S_g|},
\end{equation}

where $S_r$ and $S_g$ describe the segmented results and the actual ground truth, respectively.

After applying the Dice Index, we use the Jaccard Index, the second most frequently used image segmentation evaluation technique. The range and behavior of its value is same as that of the Dice Index. The Jaccard Index is calculated by the following:

\begin{equation}
Jaccard Index(SR,GT)= \frac{|S_r \cap S_g| }{|S_g + S_g|}
\end{equation}

Similarly, the contour matching score (e.g., the BF score) measures how close the boundary of the segmented regions matches the ground truth boundary and is calculated through the following:

\begin{equation}
BFscore(SR,GT)= \frac{\alpha_1 \alpha_2 }{\alpha_1 + \alpha_2},
\end{equation}

where $\alpha_1$ is the ratio of the number of points on the boundary of the SRs that are sufficiently close to the boundary of the GT to the length of the boundary SR, and $\alpha_2$ is the ratio of the number of points on the boundary of the GT that are sufficiently close to the boundary of the SRs to the length of the GT boundary. The range of the BF score is in [0,1]. The higher the BF score, the better the segmentation quality.

These quantitative comparisons of all models were conducted to measure the Dice Index, Jaccard Index, and BF score over the $PH^2$ [48] dataset of dermoscopic images. Three evaluation measurements are calculated for each image shown in \eqref{dermoscopic} against each model. Fig. \eqref{JI} shows the mean of these evaluation measurements in a graphical context. All image segmentation evaluation analysis techniques prove that the proposed saliency driven active contour model achieves the highest Dice Index, Jaccard Index, and contour matching score (BFscore) in comparison to the previous related models. 
\section{Discussion and Conclusion}
Region-based active contour models can be divided into global, local, and hybrid models. Global active contour models can segment images with homogeneous intensity and are independent of the initial contour position. By contrast, local active contour models enhance the efficiency of region-based models and extract complex objects having intensity inhomogeneity. Furthermore, a hybrid active contour uses the advantages of both local and global models and is successful in segmenting the required foreground from the background.

However, these models directly process the image data and are saturated to enhance efficiency. A saliency map represents image features more prominently and helps us analyze the image further. Thus, in this model, we proposed a saliency-driven image energy and used it with a pre-existing LIF model to obtain the final energy functional. Furthermore, we used regulation terms for the best fitting of the evolving curve around the boundary of the segmented object. By using the advantages of saliency with an active contour model, our model demonstrated prominent results on complex images having different intensity levels. We also conducted an experiment on different grayscale and color images as well as their datasets, to verify the model’s efficiency. The proposed model eliminated the dependency of active contour on initial contour position, noise sensitivity and image inhomgeneity. Visual representation and quantitative analysis of the segmentation results show the superiority of the proposed model on other discussed segmentation models.
\section*{Conflict of Interest}

The authors declare that there are no conflicts of interest regarding the publication of this article.

\begin{IEEEbiography}[{\includegraphics[width=1in,height=1in,clip,keepaspectratio]{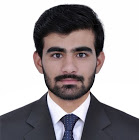}}]{Ehtesham Iqbal} received his BS degree in Electrical (computer) Engineering from COMSATS Institute of Information Technology, Pakistan, in 2017. 

He is currently working as a Research Assistant in the Visual Image media Lab  and pursuing an MS degree in the Department of Computer Science and Engineering, Chung-Ang University, South Korea. His current research interests include medical image analysis, semantic segmentation, and generative modeling. 
\end{IEEEbiography}
\begin{IEEEbiography}[{\includegraphics[width=1in,height=1in,clip,keepaspectratio]{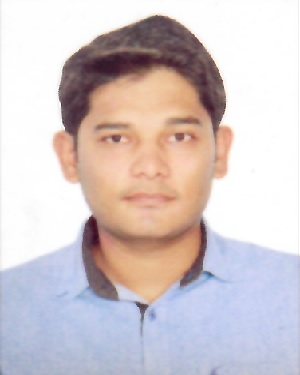}}]{Asim Niaz} received his BS degree in Electrical (computer) Engineering from COMSATS Institute of Information Technology, Pakistan, in 2016. and his MS degree from Department of Computer Science and Engineering, Chung-Ang University, South Korea, in 2020

He is a researcher with the STARS team at INRIA Sophia Antipolis France. His areas of interests are action recognition, video understanding, medical image analysis and image segmentation..
 
\end{IEEEbiography}
\begin{IEEEbiography}[{\includegraphics[width=1in,height=1in,clip,keepaspectratio]{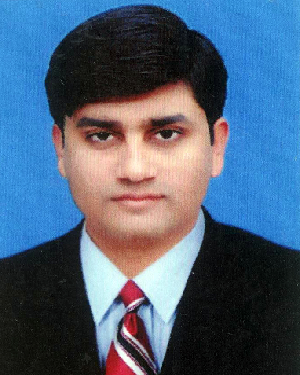}}]{Asif Aziz Memon} received his BE degree from Mehran UET, Jamshoro, Pakistan, in 2010, and his ME degree from Mehran UET, Jamshoro, Pakistan, in 2015.

He is currently pursuing a PhD degree in Application Software from Chung-Ang University, Seoul, South Korea. His research interests include image segmentation, image recognition, and medical imaging.  
\end{IEEEbiography}

\begin{IEEEbiography}[{\includegraphics[width=1in,height=1in,clip,keepaspectratio]
{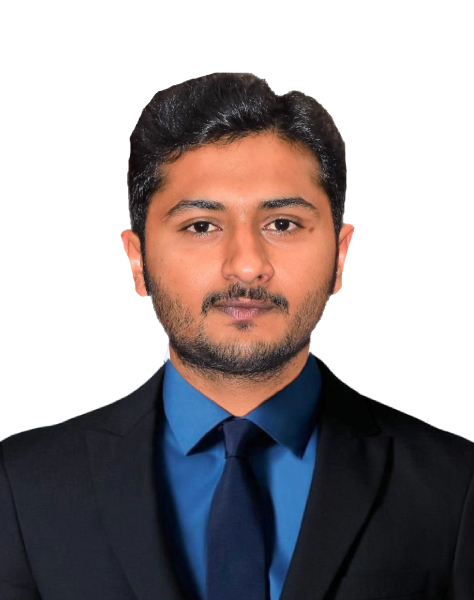}}]{Usman Asim}received the BS degree in computer science from COMSATS University Islamabad , Pakistan, in 2019.

 He is currently pursuing the M.S. degree as a graduate Research Student with the Department of Computer Science and Engineering, Chung-Ang University, Seoul, South Korea. His current research interest includes medical image analysis, semantic segmentation, and object segmentation.
\end{IEEEbiography}

\begin{IEEEbiography}[{\includegraphics[width=1in,height=1in,clip,keepaspectratio]{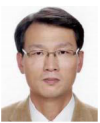}}]{Kwang Nam Choi} received his BS and MS degrees from the Department of Computer Science, Chung-Ang University, Seoul, South Korea, in 1988 and 1990, respectively, and a PhD in Computer Science from the University of York, UK, in 2002. 

He is currently a professor with the School of Computer Science and Engineering, Chung-Ang University. His current research interests include motion tracking, object categorization, and 3D image recognition.
\end{IEEEbiography}

\EOD

\end{document}